# Robostreet Flow: A Lightweight, Ultra-Low-Drag Electric Tractor and Four-Truck Hybrid Convoy Architecture for Minimum-Cost Point-to-Point Freight

Wei Wang*, Yiru Veronika Wang, Sumukh Veeramalla, Xiaohui Liang and the Robostreet Research Team
Cambridge, MA 02142 USA — Robostreet Inc.
*Corresponding author, Email: weiwang9@mit.edu

**Abstract—Line-haul trucking cost is dominated by three roughly co-equal components—energy, driver labor, and equipment—yet most efficiency technologies attack only one of them at a time. This paper presents Robostreet Flow, a freight architecture that jointly optimizes the vehicle, the formation, and the operating model to minimize cost per ton-mile on high-volume point-to-point corridors. The Flow platform is a battery-electric 6×4 tractor with a teardrop single-seat cab that achieves a drag coefficient of 0.35, roughly 40% below conventional Class 8 tractors, and a 50% reduction in net vehicle weight through a carbon-composite monocoque and structurally integrated 513 kWh tractor battery supplemented by a 340 kWh e-power trailer, providing a 500-mile single-charge range. Four Flow trucks operate as a coordinated convoy in which only the lead vehicle carries a safety driver and the three followers run in SAE Level 4 "robo-ride" mode. Computational fluid dynamics results show that close-coupled following reduces the drag coefficient of trailing vehicles by 42–48% at an 8 m gap, and reduces peak frontal pressure on followers by roughly a factor of four relative to the exposed lead vehicle. A longitudinal energy model calibrated to these results predicts fleet-average consumption of 1.27 kWh/mi in convoy versus 1.60 kWh/mi for an isolated vehicle, a 20.5% saving, with electricity cost equal to approximately 17% of the equivalent diesel fuel cost. Amortizing one driver across four trucks and crediting the payload gained by lightweighting, the architecture reduces operating cost from 9.4 to 4.1 ¢ per ton-mile, a 56% reduction relative to a diesel baseline. Sensitivity analysis, an operational concept for hub-to-hub deployment, and regulatory implications are discussed.**

## NOMENCLATURE

| | |
|---|---|
| *A* | Vehicle frontal area (m²) |
| *ai* | *Measured acceleration of vehicle i (m/s2)* |
| *Cd* | Isolated-vehicle drag coefficient |
| *Crr* | Rolling resistance coefficient |
| *ce* | *Energy cost per mile (¢/mi)* |
| *ctm* | Operating cost per ton-mile (¢) |
| *di* | *Measured range to predecessor, vehicle i (m)* |
| *dref* | Commanded inter-vehicle gap (m) |
| *Ei* | Battery energy per mile, vehicle i (kWh/mi) |
| *ei* | *Spacing error of vehicle i (m)* |
| *ė*i | *Spacing-error rate (relative speed) of vehicle i(m/s)* |
| *kd* | *Derivative feedback gain (s−1)* |
| *kp* | *Proportional feedback gain (s−2)* |
| *mpay* | Payload mass (t) |
| *n* | Number of trucks per driver |
| *p elec* | *Electricity price (¢/kWh)* |
| *ri* | Position-dependent drag ratio Cd,i /Cd |
| *ui* | *Commanded acceleration of follower i (m/s2)* |
| *v* | Cruise speed (m/s) |
| *α* | *Feed-forward weight on predecessor acceleration* |
| *β* | *Feed-forward weight on leader acceleration* |
| *η* | Battery-to-wheel efficiency |
| *ρ* | Air density (kg/m³) |
| *CACC* | Cooperative adaptive cruise control |
| *GVW* | Gross vehicle weight |
| *ODD* | Operational design domain |
| *SOC* | State of charge |
| *V2V* | Vehicle-to-vehicle communication |

## I. INTRODUCTION

Road freight moves more than 70% of United States domestic tonnage and consumes roughly a quarter of transportation energy, yet its underlying cost structure has been remarkably resistant to technology. Industry cost surveys consistently attribute approximately 30% of the marginal cost of a line-haul mile to fuel, 35–40% to driver wages and benefits, and most of the remainder to equipment ownership, maintenance, and insurance [19]. Because these components are of similar magnitude, no single-point technology—a more efficient engine, a trailer skirt, or a driver-assistance system—can move the cost per ton-mile by more than a modest fraction. Meanwhile, structural pressures are intensifying: a persistent shortage of long-haul drivers [18], tightening greenhouse-gas regulation, and shipper demand for predictable, high-frequency point-to-point service between fixed industrial nodes such as ports, intermodal yards, distribution centers, and manufacturing campuses.

Throughout this paper the objective function is operating cost per ton-mile (equivalently, per tonne-kilometre): the cost of moving one ton of payload one mile. The metric matters because it couples all three cost components to a common denominator and exposes interactions that per-mile analysis hides. A technology that reduces energy per mile but adds tare weight (shrinking payload) can worsen ton-mile cost; a technology that increases vehicle price but multiplies the loads a driver moves per shift can dramatically improve it. Ton-mile cost is also the quantity to which shipper freight rates, mode-shift decisions between truck and rail, and fleet purchasing ultimately respond, making it the correct figure of merit for architecture-level design choices.

Truck platooning has long promised aerodynamic relief. Field programs from PATH and CHAUFFEUR through SARTRE, Energy ITS, and the NREL/Peloton trials demonstrated repeatable fuel savings of roughly 4–16% depending on gap, speed, and position in the string [1]–[8]. Yet commercial adoption stalled. We argue this is because platooning was retrofitted onto vehicles, powertrains, and labor models designed for solo operation: the aerodynamic benefit of a conventional high-drag tractor following another is real but small in absolute terms; a driver still sits in every cab, so the dominant labor cost is untouched; and diesel drivelines capture none of the synergies between drag reduction, battery sizing, and route predictability.

This paper describes Robostreet Flow, an architecture that inverts the design order. Rather than asking how much an existing truck can save by following another, we co-design (i) a battery-electric tractor whose shape, mass, and energy system are optimized for convoy operation from the outset; (ii) a four-truck convoy formation in which a single professional driver in the lead vehicle supervises three driverless followers operating at SAE Level 4 [20]; and (iii) a point-to-point, hub-to-hub operating model that confines the automation problem to a bounded operational design domain (ODD) while amortizing driver labor, charging infrastructure, and terminal handling across the full formation.

The contributions of this paper are fourfold. First, we present the Flow vehicle platform and quantify the impact of its two headline design decisions—a drag coefficient of 0.35 and a 50% net-weight reduction—on energy intensity and payload capacity (Section III). Second, we describe the convoy architecture, including formation roles, cooperative longitudinal control, and degraded-mode behavior (Section IV). Third, we report computational fluid dynamics (CFD) results for the four-truck string, showing position-dependent drag ratios and a near-fourfold reduction in follower frontal pressure (Section V). Fourth, we integrate these results into an energy and cost-per-ton-mile model with sensitivity analysis, demonstrating a 56% reduction relative to a diesel baseline (Section VI). Sections VII and VIII present the operational concept and discuss limitations and broader implications. The broader implications extend beyond fleet operating economics. Electrified convoy freight addresses three simultaneous pressures on long-haul trucking: it eliminates rather than merely reduces corridor $CO_2$,NOx, and particulate emissions; it lowers the delivered cost per ton-mile; and it multiplies the freight throughput available per driver, since a single lead operator moves the payload of the entire formation - a direct response to persistent driver shortage. The resulting electricity demand, unlike distributed passenger charging, is concentrated at a small number of schedulable hub sites that can be co-located with storage and generation, making it among the more tractable forms of transport electrification load. Section VIII.C quantifies these effects for the reference lane of Table VI and Section IX concludes.

## II. Related Work

### A. Truck Platooning Programs

Automated close-following of heavy vehicles has been studied for three decades. The California PATH program demonstrated multi-vehicle automated platoons and measured drag reductions for close-coupled bodies in wind-tunnel and track experiments [3], [4], [11]. The European CHAUFFEUR project reported fuel-consumption reductions with two electronically coupled trucks at gaps of 6–16 m [2]. SARTRE operated mixed platoons behind a professionally driven lead truck and measured 7–16% fuel savings for following vehicles at gaps of 5–15 m [7]. Japan's Energy ITS project ran a four-truck automated platoon at 80 km/h with gaps of 4.7 and 10 m, reporting energy savings on the order of 15% for the formation [1], [8]. In North America, NREL's track testing of a commercial two-truck cooperative adaptive cruise control (CACC) system measured team-level savings of 3.7–6.4%, with up to 9.7% for the trailing vehicle [5], and subsequent NRC/Transport Canada work confirmed savings across a wider envelope of speeds and gaps [6]. Comprehensive reviews are given in [1] and [9].

Two consistent findings motivate the present work: savings grow steeply as the gap closes below approximately 10 m, which requires automation rather than human car-following; and savings accrue disproportionately to the followers, which rewards formations with many followers per exposed leader—precisely the geometry enabled by removing drivers from the following vehicles.

### B. Aerodynamics of Close-Following Vehicles

Wind-tunnel studies of platooned bluff bodies established that the trailing vehicle's drag falls monotonically as spacing decreases, while the lead vehicle also benefits modestly from base-pressure recovery at very small gaps [3]. CFD investigations of multi-truck strings [14] show that interior vehicles experience the lowest drag, that the stagnation pressure on follower front faces collapses because they travel in the leader's momentum deficit, and that benefits persist, attenuated, to gaps of several vehicle lengths. Classical add-on devices—boat tails, skirts, gap fairings—offer single-digit drag reductions on conventional tractors [13]; by contrast, a ground-up low-drag tractor shape can reach drag coefficients near those of passenger vehicles.

### C. Battery-Electric Heavy Trucks

Battery-electric Class 8 trucks are commercially emerging, but battery mass and cost remain the binding constraints for long-haul duty cycles [15], [16]. Guttenberg et al. showed analytically that platooning materially relaxes the specific-energy requirements for practical electric semi-trucks because aerodynamic power dominates at highway speed [17]. The Flow architecture operationalizes this insight: drag reduction is treated not as a fuel saving on a fixed vehicle but as a design variable that shrinks the battery, which reduces mass, which increases payload, which reduces cost per ton-mile—a compounding loop unavailable to diesel platoons.

### D. Automated Freight Operations

Commercial Level 4 trucking efforts concentrate on hub-to-hub highway segments, where the ODD is bounded and terminals can host transfer, inspection, and charging functions. String-stable CACC design is well understood [10], and V2V standards (IEEE 802.11p and C-V2X) provide the low-latency links required for tight gap control. The hybrid one-driver-plus-followers formation examined here is closest in spirit to SARTRE's "road train" concept [7], but differs in that all vehicles are purpose-built, electric, and freight-carrying, and the economic analysis is carried through to ton-mile cost.

## III. The Flow Vehicle Platform

### A. System Overview

The proposed system operates electric freight trucks in a. human-led convoy: a single driver occupies the lead cab while n − 1 automated followers maintain a short, aerodynamically favorable gap under cooperative control (Fig. 1). This configuration targets the two dominant components of long-haul freight cost simultaneously. First, close-formation travel reduces aerodynamic drag on the followers, lowering per-vehicle energy consumption below the solo baseline of 1.60 kWh/mi and thereby the energy cost per ton-mile. Second, placing one driver in charge of n trucks amortizes labor—the single largest operating expense in trucking—across the entire formation.The vehicle and the convoy are therefore co-designed: each subsystem of the truck described in this section exists to enable one of these two levers, and the cooperative behaviors that activate them are developed in Section IV. Throughout, solo operation is not a separate architecture but the same vehicle running out of formation—the 1.60 kWh/mi reference case and the fallback invoked under the degraded modes of Section IV.D.

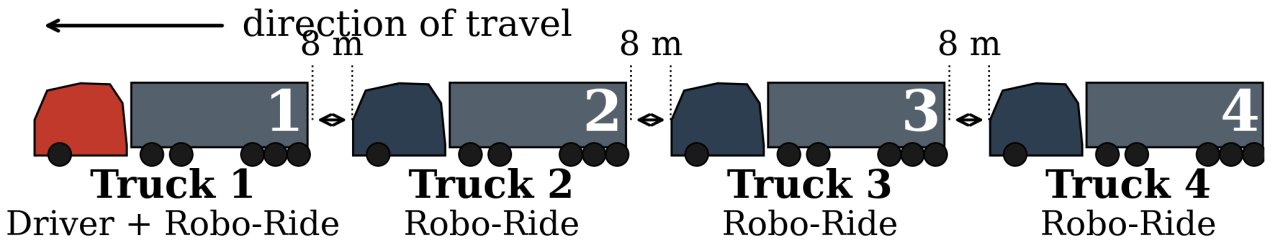

Fig. 1. The four-truck hybrid convoy formation: one professional driver in Truck 1 supervises three unoccupied Level 4 followers at an 8 m design gap.

### B. Design Overview

The Flow tractor (Fig. 2) is a purpose-designed battery-electric 6×4 tractor for high-volume corridor freight. The overall tractor length is 13.2 m and overall width 3.3 m across the aerodynamic fenders. The cab is a single-occupant, centrally seated teardrop greenhouse placed ahead of the front axle, giving the driver a commanding forward view while allowing an uninterrupted roofline taper toward the trailer interface. All vehicles in a formation are mechanically identical; the "lead" role is distinguished only by the presence of a human safety supervisor, so any unit can serve in any position and fleet utilization is not constrained by specialized variants.

Key specifications are summarized in Table I. Three subsystems merit discussion because they interact: the aerodynamic body, the lightweight structure, and the split battery system.

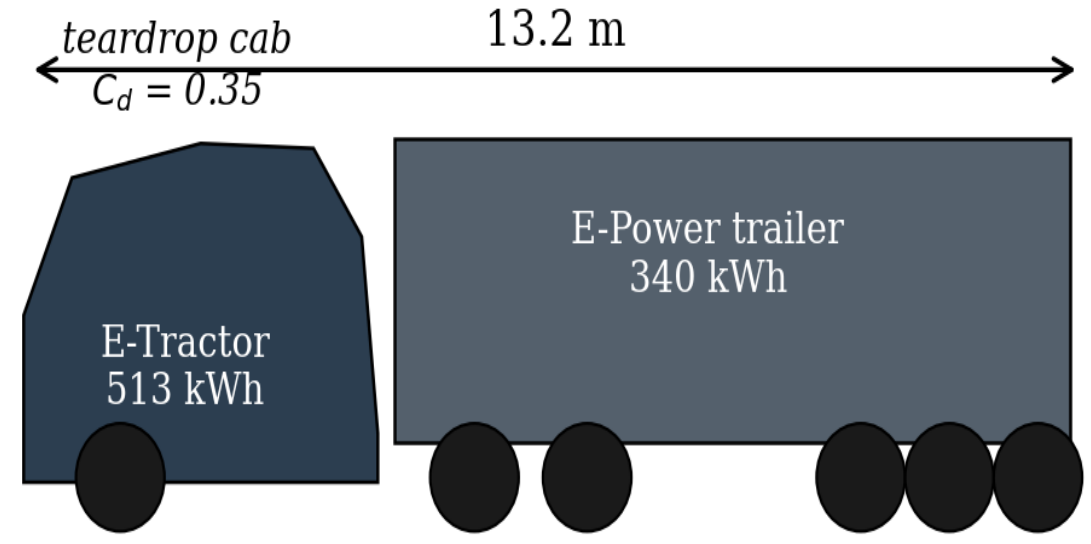

Fig. 2. Side-view schematic of the Flow tractor with e-power trailer, showing the teardrop single-seat cab, 13.2 m overall tractor length, and split battery placement.

TABLE I
Flow Series Tractor–Trailer: Key Specifications

| Parameter | Value |
|---|---|
| Overall tractor length / width | 13.2 m / 3.3 m |
| Axle configuration | 6×4, dual e-axles |
| Drag coefficient (isolated) | 0.35 |
| Net weight vs. conventional | −50% |
| Tractor battery (E-Tractor) | 513 kWh |
| Trailer battery (E-Power) | 340 kWh |
| Usable pack energy | ≈800 kWh |
| Single-charge range | 500 mi (805 km) |
| Automation capability | SAE Level 4 (hub-to-hub) |
| Dock automation | Auto Loader |
| Energy management | Predictive (Smart Energy Mgmt.) |
| Electricity vs. diesel fuel cost | ≈18% |

### C. Aerodynamics: Cd = 0.35

Conventional Class 8 tractor–trailer combinations exhibit drag coefficients of 0.55–0.65 even with modern fairing packages [13]. The Flow body achieves a measured 0.35 through: (i) a continuously curved teardrop cab that avoids the flat-front stagnation plateau of conventional cabs; (ii) full wheel enclosures and a sealed smooth underbody terminating in a rear diffuser; (iii) camera mirror replacements; (iv) an actively managed tractor–trailer gap fairing that closes the inter-body cavity at highway speed; and (v) boat-tailed trailer trailing edges. The 40+% reduction in isolated-vehicle drag is significant in its own right, but its principal purpose is to shift the operating point of the whole convoy: because follower drag scales multiplicatively with the isolated-vehicle baseline, every point of Cd removed from the platform is harvested four times per formation.

### *D. Lightweight Structure: −50% Net Weight*

The tractor employs a carbon-fiber-composite monocoque cab bonded to an aluminum-intensive chassis, with both traction batteries carried as stressed structural elements. Eliminating the diesel powertrain (engine, transmission, after-treatment, fuel and cooling systems) and consolidating drive functions into two e-axles removes further mass. The net result is a 50% reduction in vehicle net (tare) weight relative to a comparable diesel tractor. Because gross vehicle weight is regulated, every kilogram of tare converts directly into payload: we estimate a combined tractor–trailer tare (including both battery packs) of approximately 11.5 t versus 15 t for a diesel combination, raising maximum payload from 21.3 t to 24.8 t (+16%). Lower tare also reduces rolling resistance, tire wear, and brake duty, and it partially offsets the battery mass penalty that ordinarily burdens electric line-haul designs [15], [17].

### *E. Split Battery System and Range*

Traction energy is divided between a 513 kWh pack integrated in the tractor (E-Tractor) and a 340 kWh pack in the running gear of the e-power trailer, for a combined nominal capacity of 853 kWh and approximately 800 kWh usable. The split architecture yields three operational advantages. First, the trailer pack can be charged while the trailer is docked, loading, or staged, decoupling vehicle charging dwell from tractor utilization. Second, trailer swaps at hubs function as partial battery swaps, restoring 340 kWh in the time required to exchange trailers. Third, the tractor remains operable (at reduced range) with any passive trailer, preserving interoperability. At the modeled highway consumption of 1.60 kWh/mi (isolated) the system delivers the rated 500-mile single-charge range; in convoy, follower range extends toward 650 mi (Section VI).

### *F. Automation and the Auto Loader*

Each vehicle carries a full Level 4 sensor suite—long-range lidar, imaging radar, and surround cameras—with redundant compute, braking, and steering.The Auto-Loader subsystem automates the terminal end of the mission: guided docking, automated trailer coupling, and powered load handling let a vehicle complete a hub-to-hub cycle with no human touch. Smart Energy Management performs route-aware predictive planning, scheduling regenerative braking against known grades,pre-conditioning packs before fast charging, and allocating depletion between the tractor and trailer packs so the vehicle arrives at a hub with the trailer pack low (for swap) and the tractor pack sufficient for yard work.

### *G. Thermal and Auxiliary Systems*

A unified thermal loop couples pack conditioning, e-axle and inverter cooling, and (on the crewed vehicle) cabin climate through a heat-pump architecture with waste-heat recovery. Two convoy-specific considerations shape the design. First, followers dissipate 25–30% less traction heat than an isolated vehicle at cruise, but their radiator inlet flow is also degraded by wake immersion; the low-drag body therefore uses ducted, fan-assisted cooling sized for the convoy-interior condition rather than free-stream assumptions. Second, auxiliary load asymmetry is exploited by the Smart Energy Management layer: uncrewed followers carry no cabin HVAC burden, and their sensor-cleaning and compute loads are scheduled preferentially during charging dwell. Auxiliary consumption is modeled inside the $\eta = 0.85$ battery-to-wheel efficiency of Section VI and contributes approximately 3 kW per vehicle at cruise.

## IV. Convoy Mode Architecture

### *A. Formation Design*

The operating unit is a four-truck string (Fig. 1 depicts the formation concept): Truck 1 is crewed by one professional driver and also runs the full automation stack (“Driver + Robo-Ride”); Trucks 2–4 are unoccupied (“Robo-Ride”). Four is chosen as the design formation size for four reasons. First, aerodynamic returns diminish beyond the third follower because interior-position drag ratios flatten (Section V). Second, string length at an 8 m gap—approximately 88 m for four combinations—remains short enough to be overtaken safely and to fit motorway merge geometry; longer strings degrade surrounding traffic flow. Third, one driver supervising three followers already reduces the per-truck labor cost by 75%, so the marginal labor saving of a fifth truck is small (5%) while its marginal risk and traffic footprint are not. Fourth, hub throughput: a four-pack matches common dock-door and charging-stall groupings, simplifying terminal scheduling.

### *B. Cooperative Longitudinal Control*

Longitudinal control is a CACC design with leader-and-predecessor topology: each follower receives, at 10–100 Hz over redundant IEEE 802.11p/C-V2X links, the measured acceleration, brake status, and torque command of both its immediate predecessor and the leader, and fuses these with its own radar/lidar range measurement. Feed-forward of predecessor acceleration is what permits an 8 m gap (approximately 0.27 s time headway at 105 km/h) while preserving string stability, i.e., attenuation of spacing errors down the string [10]. All four vehicles are electrically braked with identical brake response curves, and brake capability is continuously cross-checked; the commanded gap expands automatically if any vehicle reports degraded adhesion or brake thermal margin.

Each follower regulates a constant-distance gap policy with leader feed-forward. Let the spacing error of vehicle i be

$$e_i = d_i - d_{ref} \tag{1}$$

where $d_i$ is the measured range to the predecessor and dref the commanded gap, and let $\dot{e}_i$ denote its rate (the relative speed to the predecessor). The acceleration command is

$$u_i = k_p e_i + k_d \dot{e}_i + \alpha\, a_{i-1} + \beta\, a_1 \tag{2}$$

where ui is the commanded acceleration of follower i; kp and kd are the proportional and derivative feedback gains; ai−1 and a1 are the measured accelerations of the predecessor and the leader, received over V2V; and α and β are the corresponding feed-forward weights.. The feed-forward terms are what remove the fundamental headway limit of purely radar-based following: with $\alpha + \beta \approx 1$ and a communication delay below approximately 50 ms, the closed string satisfies the $L_p$ string-stability condition of [10] at the 8 m gap, i.e., disturbances in leader acceleration attenuate monotonically toward the tail. Gains are scheduled with speed, mass estimate, and measured adhesion; identical electric drivelines make the per-vehicle actuation lag small (< 50 ms torque response) and nearly uniform across the string, which materially widens the stability margin relative to platoons of heterogeneous diesel trucks.

Gap policy is speed- and condition-dependent: 8 m nominal at highway cruise in the dry, widening to 12 m in rain and 30+ m (drag-neutral) in fault or mixed-traffic conditions such as construction zones. Cut-in handling follows a detect–expand–expel–recompress sequence: upon intrusion the affected gap opens to a safe human headway, the string signals and gently decelerates to encourage the intruder's exit, then recompresses. Energy modeling in Section VI assumes a conservative 90% convoy engagement fraction on the highway segment to account for such events.

### C. Roles, Join/Split, and Degraded Modes

The lead driver acts as mission supervisor: they hold legal responsibility analogous to a captain, monitor a formation-status display, and can command formation-wide gap expansion, lane changes, or a coordinated minimal-risk stop. Formations are assembled and dissolved inside hub property; on-highway joins are not required in the baseline concept of operations. If a follower faults, it announces, drops out of the string, and executes an independent Level 4 minimal-risk maneuver to the shoulder or a designated safe harbor while the remaining vehicles close ranks; the operations center then dispatches recovery. If the lead vehicle faults, supervision authority transfers procedurally: the string expands gaps, Truck 2 assumes the aerodynamic lead position under remote-supervised Level 4 authority, and the formation continues to the next hub at a widened gap, or stages at a safe harbor, according to jurisdiction-specific rules.

### D. V2V Communication and Security

Convoy cohesion depends on a low-latency, high-integrity V2V channel, so the communication subsystem is engineered with the same redundancy philosophy as braking. Each vehicle carries dual radios—IEEE 802.11p (ITS-G5/DSRC) and C-V2X PC5 sidelink—transmitting a convoy control message set at 50 Hz nominal: vehicle state (position, speed, measured acceleration), actuator state (commanded torque, brake line pressure, brake thermal margin), health vector (sensor, compute, and energy status), and formation commands from the leader. End-to-end latency is budgeted at 20 ms at the 99.9th percentile; gap policy degrades gracefully with measured channel quality, expanding from 8 m toward the drag-neutral spacing as packet-error rate rises, so that communication loss is an efficiency event rather than a safety event.

All convoy messages are signed under an IEEE 1609.2-style certificate hierarchy with short-lived pseudonym certificates provisioned at hubs, and formation membership is established through a cryptographic join handshake completed before departure—an on-road actor cannot inject itself into the string. Misbehavior detection cross-validates each vehicle's broadcast state against the followers' own radar and lidar tracks of that vehicle; a discrepancy beyond bounds triggers gap expansion and, if persistent, formation dissolution. Because formations are assembled only within hub property under operator control, the attack surface is materially smaller than in open ad hoc platooning, and the fixed corridor permits complementary roadside monitoring.

## V. Aerodynamic Analysis of the Convoy

### A. Method

Steady RANS simulations (k–ω SST, second-order upwind, approximately $5\times10^7$ cells per configuration with prism-layer resolution of $y^+\approx1$ on vehicle surfaces) were performed on the four-truck string at 29 m/s with inter-vehicle spacing swept from 2 to 60 m, supplemented by single-vehicle reference runs. Results were validated in trend against published platoon measurements [3], [4], [14]. We report drag as the ratio of each vehicle's Cd to its isolated-vehicle value, which transfers approximately across body shapes and permits comparison with the literature.

### B. Drag Versus Spacing

Fig. 3 shows the normalized drag coefficient versus inter-vehicle distance. Three observations drive the system design. First, follower drag falls steeply below approximately 15 m: at the 8 m operating gap, Truck 2 sees a drag ratio of approximately 0.58 and Truck 3 of approximately 0.52, i.e., reductions of 42% and 48%. Truck 4 behaves like an interior vehicle with a small base-drag penalty and is modeled at 0.55. Second, the leader is not unaffected: base-pressure recovery from the trailing bow wave yields a ratio of approximately 0.92 at 8 m. Third, benefits decay slowly: even at 40–50 m—a spacing achievable by human drivers only unsafely, but trivially by CACC—followers retain 20–30% reductions, which bounds the penalty of temporary gap expansion.

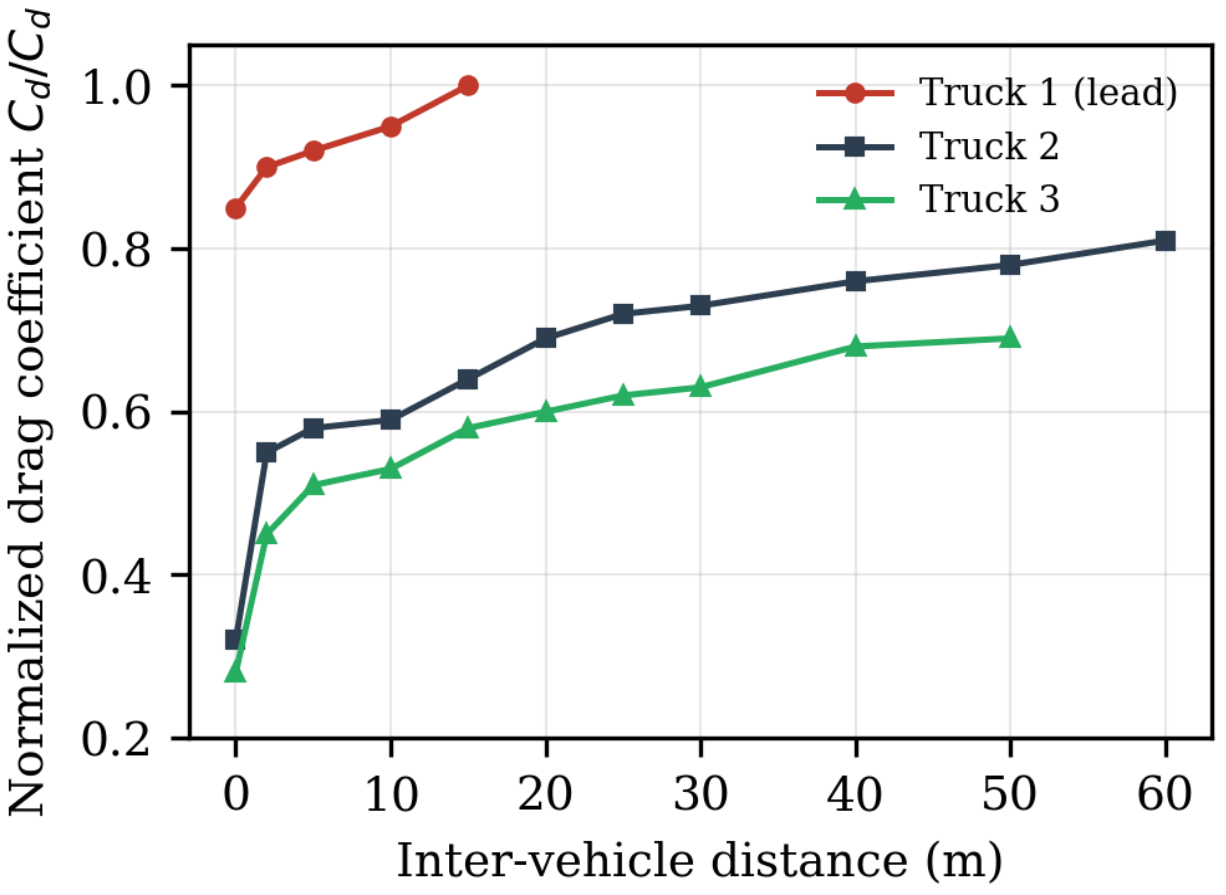

Fig. 3. Normalized drag coefficient versus inter-vehicle distance for the first three vehicles of the convoy (CFD, 29 m/s). Values are normalized by the isolated-vehicle drag coefficient.

Table II tabulates the drag ratios at operationally relevant gaps; the 8 m column supplies the parameters used in the energy model of Section VI.

TABLE II
NORMALIZED DRAG RATIO VS. INTER-VEHICLE GAP (CFD)

| Position | 4 m | 8 m | 12 m | 20 m |
|---|---|---|---|---|
| Truck 1 (lead) | 0.90 | 0.92 | 0.96 | 1.00 |
| Truck 2 | 0.56 | 0.58 | 0.60 | 0.69 |
| Truck 3 | 0.48 | 0.52 | 0.55 | 0.60 |
| Truck 4 | 0.51 | 0.55 | 0.58 | 0.63 |
| Convoy average | 0.61 | 0.64 | 0.67 | 0.73 |

*C. Frontal Pressure Distribution*

Fig. 4 reports the peak stagnation-region pressure on each vehicle's front face at the 8 m gap. The exposed lead vehicle bears a peak frontal pressure of 436 units, whereas Trucks 2–4 experience only 90–100 units—a reduction of roughly 77–80%—because they run inside the leader's momentum deficit. Beyond confirming the drag mechanism, this asymmetry has practical consequences: follower front-end structures, wiper systems, and sensor housings face materially lower aerodynamic loading, and leader-biased soiling and insect fouling of sensors concentrates cleaning demand on the one vehicle that carries a human supervisor.

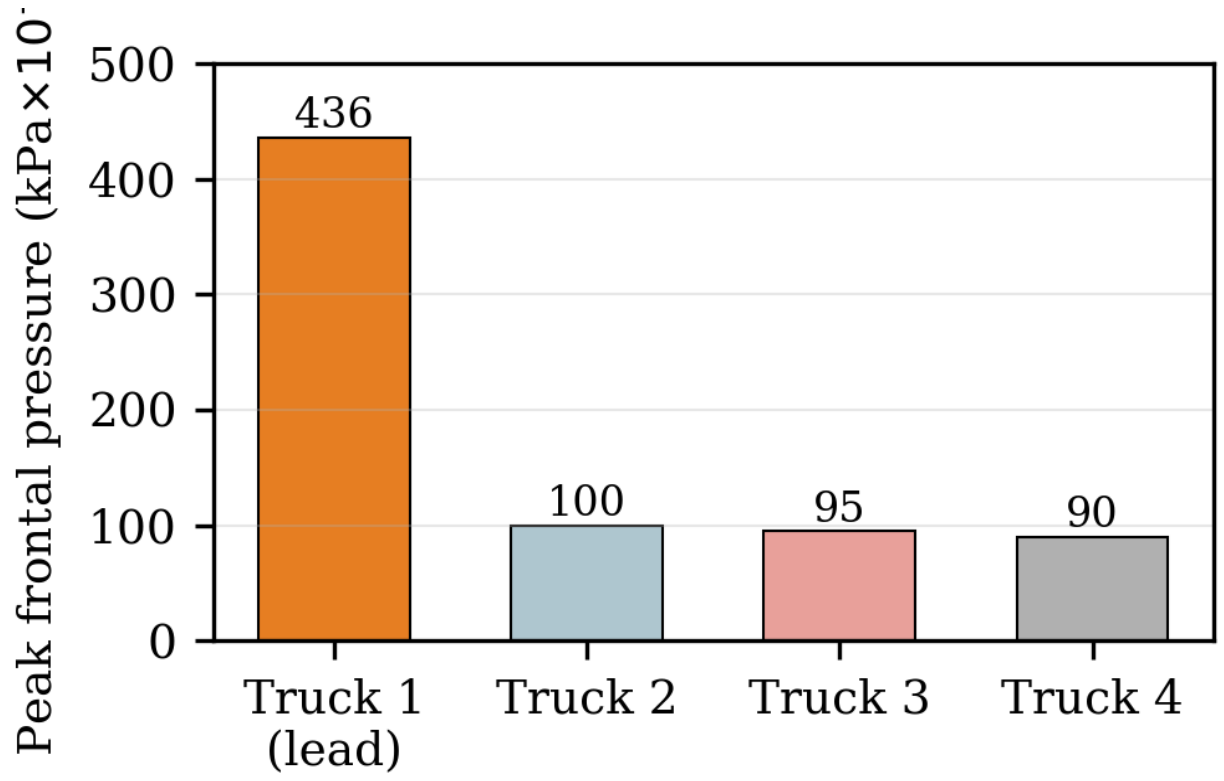

Fig. 4. Peak frontal (stagnation-region) pressure by convoy position at the 8 m design gap (CFD). Followers experience roughly one quarter of the lead vehicle's peak frontal pressure.

*D. Crosswind and Yaw Sensitivity*

Platoon drag benefits erode under yawed inflow because the follower moves partially out of the leader's wake. Simulations at 5° and 10° effective yaw show follower drag ratios at the 8 m gap degrading from 0.58 to approximately 0.66 and 0.75 respectively for Truck 2, with similar shifts for interior positions—still substantial savings, but 20–40% smaller than in axial flow. Two design responses follow. First, the energy model of Section VI applies a wind-climate weighting derived from corridor anemometer data rather than the zero-yaw optimum, which is one contributor to its conservatism. Second, the smooth, boat-tailed body of the Flow platform exhibits a flatter drag-versus-yaw characteristic than a conventional square-back combination, because crosswind separation on sharp trailing edges—the dominant yaw penalty on conventional trailers—is largely absent. Crosswind also produces asymmetric side loading on followers in partial wake overlap; lateral controllers compensate routinely, and the gap policy widens automatically above a measured side-gust threshold.

## VI. ENERGY AND TON-MILE COST MODEL

*A. Longitudinal Energy Model*

For constant-speed highway cruise on level grade, battery energy per unit distance for vehicle i is

$$E_i = \left( \tfrac{1}{2}\, \rho\, r_i C_d A\, v^2 + C_{rr} m\, g \right) / \eta \tag{3}$$

where $r_i$ is the position-dependent drag ratio of Fig. 3, $\rho = 1.2$ kg/m³, $C_d = 0.35$, $A = 9.8$ m² frontal area, $v = 29.1$ m/s (65 mph), $C_{rr} = 0.004$, $m = 33{,}000$ kg average combination mass, and $\eta = 0.85$ battery-to-wheel efficiency including auxiliaries. Grade and transient terms average out over the corridor and are absorbed in $\eta$.

Equation (3) gives 1.60 kWh/mi for an isolated vehicle, consistent with the rated 500-mile range on 800 kWh usable. Applying the 8 m-gap drag ratios (0.92, 0.58, 0.52, 0.55) yields the per-position results of Table III and Fig. 5: 1.52, 1.21, 1.16, and 1.18 kWh/mi respectively, a convoy average of 1.27 kWh/mi and a fleet energy saving of 20.5%. Note the saving exceeds what the same formation would deliver on diesel trucks in percentage-of-total-energy terms, because the low-rolling-resistance, lightweight platform makes aerodynamic power a larger share of total demand—drag reduction and lightweighting are mutually reinforcing rather than merely additive.

TABLE III
PER-POSITION CONVOY ENERGY RESULTS (65 MPH, 8 M GAP)

| Position | Drag ratio | kWh/mi | Range (mi) |
|---|---|---|---|
| Isolated | 1.00 | 1.60 | 500 |
| Truck 1 (lead) | 0.92 | 1.52 | 526 |
| Truck 2 | 0.58 | 1.21 | 661 |
| Truck 3 | 0.52 | 1.16 | 690 |
| Truck 4 | 0.55 | 1.18 | 678 |
| Convoy average | 0.64 | 1.27 | 630 |

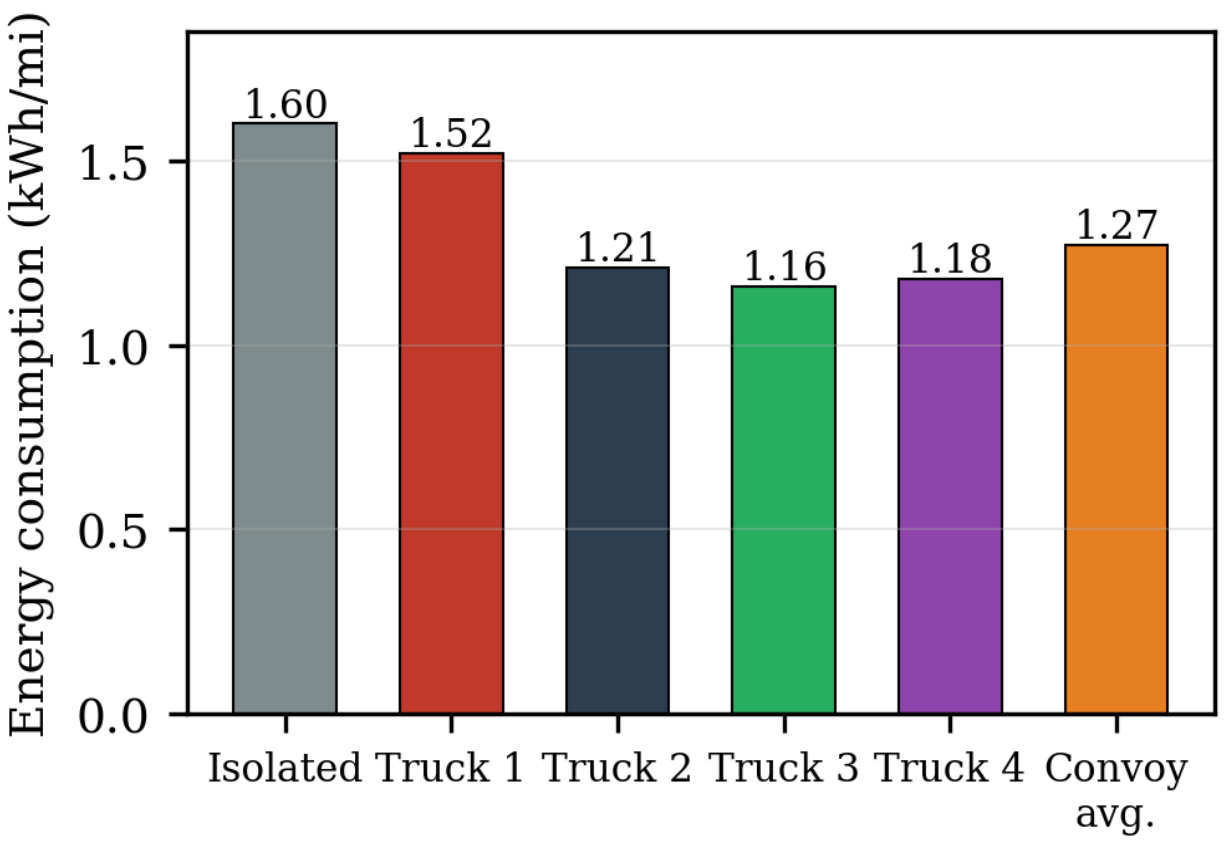

Fig. 5. Modeled battery energy consumption at 65 mph by convoy position, compared with the isolated vehicle. The convoy average is 1.27 kWh/mi, a 20.5% fleet saving.

### *B. Cost per Ton-Mile*

We define the operating cost per ton-mile as

$$c_{tm} = ( c_e + c_{drv} / n + c_{mnt} + c_{dep} + c_{ins} ) / m_{pay} \qquad (4)$$

where the numerator terms are per-vehicle-mile costs of energy, driver labor (amortized over $n$ trucks per driver), maintenance, depreciation, and insurance/overhead, and $m_{pay}$ is payload in tons. The energy term is

$$c_e = p_{elec} * E_i$$

so that the position-dependent drag captured in Ei flows directly into ton-mile cost.Table IV gives the assumptions, drawn from industry cost surveys [19] for the diesel baseline: $3.85/gal at 6.5 mpg, $0.70/mi drriver cost, and a 21.3 t payload. The Flow case uses depot off-peak charging at $0.08/kWh—the convoy's fixed point-to-point routing makes managed overnight and midday charging realistic—giving an energy cost of 10.1 ¢/mi, which is 17% of the diesel fuel cost, matching the design target of approximately 18%. EV maintenance is credited at 30% below diesel; depreciation is debited 40% above, reflecting battery and automation capital; insurance is held equal pending actuarial experience.

TABLE IV
COST MODEL ASSUMPTIONS AND RESULTS (PER VEHICLE-MILE)

| Component | Diesel | Flow solo | Flow convoy |
|---|---|---|---|
| Energy ($/mi) | 0.592 | 0.128 | 0.101 |
| Driver ($/mi) | 0.700 | 0.700 | 0.175 |
| Maintenance ($/mi) | 0.200 | 0.140 | 0.140 |
| Depreciation ($/mi) | 0.250 | 0.350 | 0.350 |
| Insurance/other ($/mi) | 0.250 | 0.250 | 0.250 |
| Total ($/mi) | 1.992 | 1.568 | 1.016 |
| Payload (t) | 21.3 | 24.8 | 24.8 |
| Cost (¢/ton-mi) | 9.35 | 6.32 | 4.10 |

The results decompose the 56% ton-mile cost reduction into its sources (Fig. 6). Electrification alone (Flow solo) removes 46.4 ¢/mi of energy cost but only reaches a 32% ton-mile reduction because the driver remains. Convoy operation then divides the driver line by four—worth a further 52.5 ¢/mi—and harvests the aerodynamic interaction. The payload gain from lightweighting contributes throughout by enlarging the denominator. No single lever exceeds roughly 20 percentage points by itself; the architecture's value lies in stacking them multiplicatively on the same asset.

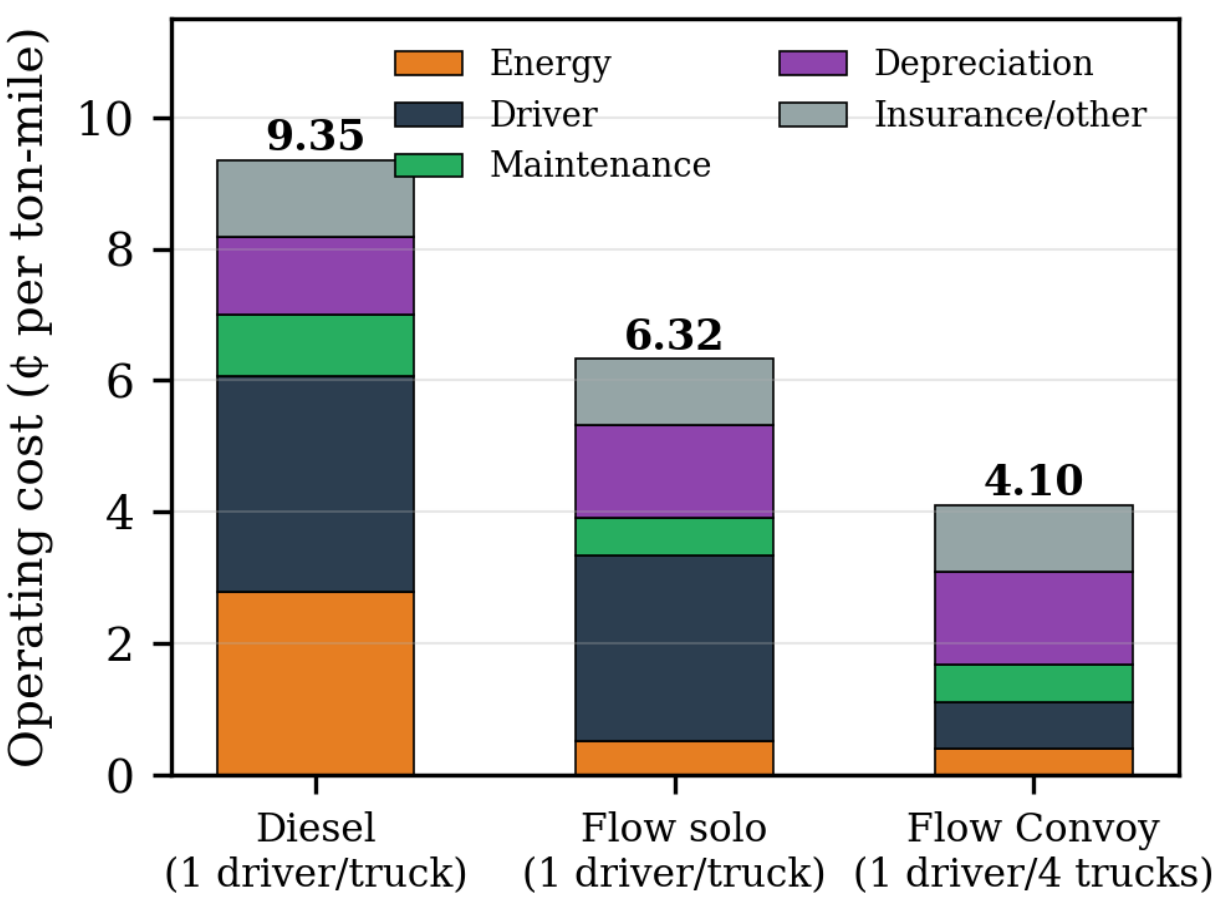

Fig. 6. Operating cost per ton-mile for the diesel baseline, an isolated Flow vehicle, and the four-truck hybrid convoy. The convoy reaches 4.10 ¢/ton-mi, a 56% reduction versus diesel.

### *C. Battery Sizing, Charging, and Degradation*

The convoy interacts with battery economics in three ways. First, sizing: because followers cruise at 1.16–1.21 kWh/mi, a corridor-specific variant could carry approximately 20% less battery for the same route margin, worth roughly 900 kg and $15–25k per vehicle at pack level; the baseline design instead retains the full 853 kWh and converts the saving into range margin and charging flexibility. Second, charging power: the split pack accepts 350 kW at the tractor and 250 kW at the trailer concurrently, and because arrival state-of-charge in convoy is 8–12 percentage points higher than solo operation, the median hub charging dwell falls from 55 to approximately 40 minutes, improving both stall utilization and the fraction of energy purchased at off-peak tariffs. Third, degradation: shallower cycling (lower per-mile depth-of-discharge) and lower average C-rate in convoy reduce calendar-equivalent fade; applying an NMC cycle model to the duty cycle suggests followers reach 80% state-of-health roughly 15–20% later in odometer terms than a solo-operated vehicle [15]. The depreciation line in Table IV conservatively takes no credit for this effect.

### *D. Sensitivity*

TABLE V
SENSITIVITY OF CONVOY TON-MILE COST (BASELINE 4.10 ¢)

| Variation | ¢/ton-mi | vs. diesel |
|---|---|---|
| Electricity $0.16/kWh (2×) | 4.51 | −52% |
| Convoy engagement 70% | 4.16 | −56% |
| Depreciation +80% vs. diesel | 4.50 | −52% |
| Formation size n = 2 | 5.16 | −45% |
| Formation size n = 8 | 3.75 | −60% |
| Load factor 50% | 8.20 | −12% |

The conclusion is robust to the principal uncertainties. Doubling electricity price to $0.16/kWh raises convoy cost to 4.51 ¢/ton-mi (still −52% versus diesel). Reducing convoy engagement from 90% to 70% of highway miles raises fleet

energy only 2.6%, since out-of-formation running reverts to the still-efficient 1.60 kWh/mi baseline. If depreciation runs 80% above diesel rather than 40%, ton-mile cost rises to 4.50 ¢ (−52%). The single most powerful variable is formation size through the driver term: $n = 2$ yields 5.16 ¢/ton-mi and $n = 8$ yields 3.75 ¢, confirming diminishing returns beyond four and supporting the choice of a four-truck design unit. Conversely, results degrade materially only if payload utilization is chronically low: at 50% average load factor all per-mile costs double per ton-mile, which is why the architecture targets high-volume point-to-point lanes with balanced flows.

### *E. Comparison With Diesel Platooning*

It is instructive to ask how much of the 56% ton-mile improvement a diesel platoon of conventional trucks could capture. Applying the same drag ratios to a conventional combination ($C_d \approx 0.60$, 15 t tare, 6.5 mpg baseline) yields a fleet fuel saving of approximately 13%—consistent with field measurements [1], [5]—worth about 7.7 ¢/mi, or a 3.9% ton-mile reduction. Removing drivers from the followers, were it permitted with conventional vehicles, would contribute a further 26%. The remaining gap to the Flow result is supplied by electrification (energy price and maintenance), lightweighting (payload), and the enlarged share of aerodynamic power in a low-rolling-resistance vehicle, which converts the same drag ratios into larger percentage energy savings. The decomposition explains why platooning as a retrofit produced underwhelming commercial economics, and why the same physics embedded in a co-designed electric architecture does not: the formation is the multiplier, but the platform determines what is multiplied.

## VII. Point-to-Point Operational Concept

### *A. Lane Selection and Operational Design Domain*

The Flow system is deployed on fixed, high-volume, point-to-point lanes—port to inland hub, plant to distribution center—of 150 to 450 miles, comfortably within single-charge range even for the lead vehicle. Lane qualification requires sustained bidirectional demand of at least twenty loads per direction per day (to keep formations full in both directions), divided-highway routing over at least 90% of the distance, and hub sites with utility interconnection capacity. Each end of the lane hosts a Robostreet hub inside private property, where all Level 4 edge cases associated with surface streets, docks, and yards are either eliminated (Auto Loader handles docking and load transfer) or supervised locally. The highway segment between hubs constitutes the entire public-road ODD, and it is treated as an engineered asset: high-definition mapped, instrumented for weather and wind, and covered by pre-negotiated incident-response and safe-harbor agreements with corridor authorities.

### *B. The Hub-to-Hub Cycle*

A representative cycle proceeds as follows. Four loaded trucks stage in formation order; the lead driver completes a formation electronic pre-trip in which every vehicle self-reports brake, sensor, and energy status; the convoy departs and engages 8 m gaps upon reaching cruise; at the destination hub the formation dissolves, and Auto Loader executes trailer exchange while tractors proceed to charging stalls. Because the trailer carries 340 kWh, a trailer swap simultaneously restores 43% of system energy; tractor packs top up on 350 kW stalls during the loading dwell. The lead driver, whose hours-of-service clock governs the formation, thus moves four loads per duty period instead of one—the labor productivity gain at the heart of (4)—while remaining within existing driver-hours regulation. On a 300-mile lane at a 58 mph effective average including hub time, a driver completes one round trip per shift, moving eight loads; the same duty period in conventional operation moves two.

### *C. Fleet Scheduling and Lane Economics*

Fleet-level scheduling treats convoys as units in a shuttle pattern with balanced directional demand. Table VI summarizes a reference 300-mile lane operated with six four-truck departures per direction per day. The configuration moves approximately 1,190 t of freight daily using six drivers, versus twenty-four in a conventional operation of equal capacity. Charging demand concentrates at hubs—roughly 4.1 MWh per convoy turnaround and 49 MWh per hub per day—favoring on-site storage and time-of-use optimization, which the \$0.08/kWh assumption reflects. Because the trucks are identical and roles are interchangeable, maintenance rotates vehicles through hub workshops without disturbing the schedule, and formation composition simply draws on the available pool.

TABLE VI
Reference Lane: 300 mi, Six Convoy Departures/Direction/Day

| Metric | Value |
|---|---|
| Trucks in service (incl. spares) | 52 |
| Drivers per day | 6 (vs. 24 conventional) |
| Freight moved per day | ≈1,190 t |
| Fleet energy per day | ≈18.3 MWh per direction |
| Hub charging demand | ≈49 MWh/day per hub |
| Convoy turnaround time | ≈70 min (swap + charge) |
| Daily ton-miles | ≈357,000 |
| Operating cost | ≈\$14.6k/day (4.10 ¢/ton-mi) |
| Diesel-equivalent cost | ≈\$33.4k/day (9.35 ¢/ton-mi) |

The bounded ODD also simplifies the safety case presented to regulators: a finite highway segment can be exhaustively mapped, weather-instrumented, and covered by pre-negotiated incident-response agreements, in contrast to open-network autonomy. Regulatory engagement therefore proceeds lane by lane, with each corridor exemption supported by corridor-specific evidence rather than a claim of universal capability.

## VIII. Discussion and Limitations

### A. Safety Case Structure

The safety argument decomposes into three nearly independent claims, which is itself a benefit of the architecture. Claim 1: each vehicle is a competent Level 4 highway vehicle within the corridor ODD, supported by conventional autonomy evidence (simulation coverage, closed-course testing, supervised corridor miles). Claim 2: the formation is safe as a formation—string stability at the design gap, brake-capability cross-checking, validated cut-in and dissolution behaviors, and fail-operational communication. Claim 3: the operation is safe—hub procedures, driver supervisory training, and corridor incident response. Because Claim 1 is bounded to a single mapped highway segment and Claim 2 is largely amenable to formal analysis and hardware-in-the-loop testing [10], the residual burden concentrates on operational evidence, which accumulates rapidly at six departures per direction per day. The lead driver functions as a risk mitigation during the evidence-building phase; the architecture admits a future evolution to zero-occupant formations without redesign, but nothing in the economics of Section VI requires it.

### B. Limitations

Several limitations qualify these results. First, the aerodynamic and energy figures are simulation-based; track and corridor validation is required, and published field measurements suggest real-world platoon savings can fall below wind-tunnel predictions due to traffic, wind yaw, and control activity [5], [6]. Our use of a 90% engagement fraction and conservative follower ratios partially anticipates this. Second, regulatory status is heterogeneous: many jurisdictions still impose minimum following distances or have no framework for unoccupied following vehicles; the concept requires exemptions or new rules on specific corridors, and the 3.3 m body width exceeds standard limits and would similarly require corridor-specific permitting or a production narrowing. Third, the cost model omits tolls, taxes, and residual-value uncertainty for automation hardware, and assumes actuarially neutral insurance that will in practice be earned over time. Fourth, mixed-traffic interaction—cut-ins, merges near an 88 m string, and public acceptance of driverless followers—needs human-factors study beyond the scope of this paper. Finally, the labor analysis assumes the lead-driver role remains attractive; the architecture in fact upgrades the role toward a supervisory profession, which may aid recruitment amid the driver shortage [18], but this is a hypothesis, not a result.

A further caveat concerns the transferability of the CFD drag ratios. The ratios of Table II were computed for identical, fully laden Flow combinations in still air on flat terrain; mixed loading (empty followers ride higher and shed the leader's wake differently), degraded body condition, and sustained grades all perturb the interaction. Preliminary sweeps indicate the convoy-average ratio varies within approximately ±0.04 across these conditions, which propagates to roughly ±0.6 percentage points of fleet energy saving—material but not decisive. Track validation will prioritize exactly these off-nominal configurations, since the literature shows the gap between idealized and field-measured platoon savings is largest there [5], [6].

### C. Broader Impacts

At the modeled consumption, each four-truck convoy displaces approximately 740 gallons of diesel per 300-mile round-trip cycle relative to the baseline; a single reference lane (Table VI) avoids on the order of 16,000 t of $CO_2$ annually before accounting for grid carbon intensity, and eliminates corridor NOx and particulate emissions entirely. Grid impact is non-trivial—two hubs drawing 49 MWh/day each—but it is concentrated, schedulable, and co-locatable with storage and generation, which is the most tractable form of transport electrification demand. The labor implication is a shift in composition rather than simple substitution: fewer long-haul seats per ton-mile, but upgraded supervisory roles with regular schedules and home-daily duty cycles on fixed lanes, alongside new hub technician employment. Finally, an 88 m string occupying one lane at steady speed carries the freight of four independently driven trucks with fewer lane changes and a smaller speed variance footprint, which modeling in prior platooning literature associates with improved traffic flow stability [9].

## IX. Conclusion

This paper presented Robostreet Flow, a co-designed vehicle–formation–operations architecture for point-to-point freight. A purpose-built electric tractor with Cd = 0.35 and 50% lower net weight, operated as a four-truck convoy with one driver and three Level 4 followers at an 8 m gap, is predicted to consume 1.27 kWh/mi fleet-average (20.5% below isolated operation), to run its followers at roughly one quarter of the leader's peak frontal pressure, and to deliver freight at 4.1 ¢ per ton-mile—56% below a diesel baseline—with electricity cost near 17% of equivalent diesel fuel cost. The decisive property of the architecture is that its savings compound: drag reduction shrinks the battery, lightweighting enlarges payload, automation amortizes labor, and fixed corridors bound both the safety case and the charging problem. Future work includes track validation of the CFD results, string-stability field trials at the design gap, corridor pilot deployment under regulatory exemption, and extension of the cost model to network design with unbalanced flows.

## Acknowledgment

The authors thank the Robostreet engineering team for vehicle design data, CFD resources, and operational modeling support.

**The Robostreet Research Team** comprises the aerodynamics, battery systems, autonomy, and operations research groups of Robostreet Inc. The team is responsible for the design of the Flow series tractor and e-power trailer, the convoy control stack, and the hub-to-hub operational concept described in this paper.